\documentclass[conference]{IEEEtran}
\IEEEoverridecommandlockouts
\usepackage{cite}
\usepackage{amsmath,amssymb,amsfonts}
\usepackage{algorithmic}
\usepackage{graphicx}
\usepackage{textcomp}
\usepackage{xcolor}
\usepackage{booktabs,tabularx}
\usepackage{siunitx}
\usepackage{todonotes}
\usepackage{caption}
\usepackage{subfig}

\usepackage{tikz}
\usepackage{hyperref}

\newcommand\copyrighttext{%
  \footnotesize \textcopyright 2020 IEEE. Personal use of this material is permitted.
  Permission from IEEE must be obtained for all other uses, in any current or future
  media, including reprinting/republishing this material for advertising or promotional
  purposes, creating new collective works, for resale or redistribution to servers or
  lists, or reuse of any copyrighted component of this work in other works.\\
  Accepted for publication in: 3RD IEEE International Conference on Industrial Cyber-Physical Systems (ICPS 2020).}
\newcommand\copyrightnotice{%
\begin{tikzpicture}[remember picture,overlay]
\node[anchor=south,yshift=10pt] at (current page.south) {\fbox{\parbox{\dimexpr\textwidth-\fboxsep-\fboxrule\relax}{\copyrighttext}}};
\end{tikzpicture}%
}

\newcolumntype{C}[1]{>{\centering\let\newline\\\arraybackslash\hspace{0pt}}m{#1}}

\def\BibTeX{{\rm B\kern-.05em{\sc i\kern-.025em b}\kern-.08em
    T\kern-.1667em\lower.7ex\hbox{E}\kern-.125emX}}
\begin{document}

\title{A Feature Importance Analysis for  Soft-Sensing-Based Predictions in a Chemical Sulphonation Process\\
}

\author{\IEEEauthorblockN{Enrique Garcia-Ceja, \AA smund Hugo, \\ Brice Morin}
\IEEEauthorblockA{\textit{Software and Service Innovation} \\
\textit{SINTEF Digital}\\
Oslo, Norway \\
\{enrique.garcia-ceja, \\ aasmund.hugo,brice.morin\}@sintef.no}
\and
\IEEEauthorblockN{Per Olav Hansen, Espen Martinsen}
\IEEEauthorblockA{\textit{}
\textit{Unger Fabrikker}\\
Fredrikstad, Norway \\
\{Per.Olav.Hansen,Espen.Martinsen\}@unger.no}
\and
\IEEEauthorblockN{An Ngoc Lam, \O ystein Haugen}
\IEEEauthorblockA{\textit{Department of Informatics} \\
\textit{\O stfold University College}\\
Halden, Norway \\
\{an.n.lam,oystein.haugen\}hiof.no}
}

\maketitle
\copyrightnotice

\begin{abstract}
In this paper we present the results of a feature importance analysis of a chemical sulphonation process. The task consists of predicting the neutralization number (NT), which is a metric that characterizes the product quality of active detergents. The prediction is based on a dataset of environmental measurements, sampled from an industrial chemical process. We used a \emph{soft-sensing} approach, that is, predicting a variable of interest based on other process variables, instead of directly sensing the variable of interest. Reasons for doing so range from expensive sensory hardware to harsh environments, e.g., inside a chemical reactor. The aim of this study was to explore and detect which variables are the most relevant for predicting product quality, and to what degree of precision. We trained regression models based on linear regression, regression tree and random forest. A random forest model was used to rank the predictor variables by importance. Then, we trained the models in a forward-selection style by adding one feature at a time, starting with the most important one. Our results show that it is sufficient to use the top $3$ important variables, out of the $8$ variables, to achieve satisfactory prediction results. On the other hand, Random Forest obtained the best result when trained with all variables.
\end{abstract}


\section{Introduction}
\label{sec:introduction}

Strategies to reduce the negative influence on the environment  have recently become integral to industrial engineering, and it is likely that they will frame the field for the next years \cite{moses}, with machine learning as one of the surging techniques to improve efficiency.


During chemical production processes, sensing the main parameters can be hard, expensive or even infeasible. This can be attributed to some factors such as complex non-linear relations, harsh environments and expensive apparatus. In order to overcome some of those limitations, the idea of \emph{soft sensing} has been proposed \cite{Ge2017}. Within the soft sensing paradigm, variables that are easy to measure are used to infer the ones that are more difficult to monitor. In chemical processes, some output variables' values are calculated by means of doing offline laboratory analyses which are often time consuming, and may even induce a production halt. In such production processes, \emph{soft sensing} concepts can be advantageous. An example of this is the model based on soft sensors proposed by Geng et al. which was used to infer important variables within a Purified Terephthalic Acid (PTA) process \cite{GENG201738}. They built a soft sensor model that predicts the consumption of acetic acid based on the PTA solvent system data. Their implementation was based on a neural network.
Zhang et al. presented a similar idea but in this case applied to predict the quality of cobalt oxalate \cite{ZHANG20131267}. The automatic approach estimates the particle size which depends on different process parameters like flow rate of ammonium oxalate, the temperature of the reactor, and so on.

In our previous work \cite{gc2019}, we used soft sensing to estimate the neutralization number (NT) that characterizes the quality of the product during a sulphonation process, at a factory called Unger Fabrikker in Norway. We did so by training regression models such as linear regression, decision tree and Random Forest. We achieved good performance by utilizing all the process variables to train the models. However, two important questions were not fully addressed:
\begin{enumerate}
  \item \emph{Which are the most relevant variables when predicting the NT number?}
  \item \emph{Can models be trained accurately, using only a subset of the most important variables?}
\end{enumerate}
This is a vital question for industrial processes, since it serves as the basis for implementing more robust automated systems. For example, with regards to redundancy, one wish to know if it is possible to predict the value of interest to a satisfactory degree of precision, if certain sensors stop functioning or if certain values suffer from loss of integrity. Thus, one can build more trustworthy and robust systems knowing the implications when the values of some sensors may be missing or not reliable due to failures.

In this paper we performed a feature importance analysis that extends our previous work. In machine learning terminology, a feature is a variable from the dataset that is used as a predictor. We identified the most important features. Then, we trained the models with different subsets of input features, starting with just the most important feature and then adding feature by feature until all are covered. Once we identified the relevant features, we only used those to train predictive models and test their performance on an independent test set. Our results showed that by using the top $3$ most important variables it was possible to predict the NT value with good accuracy.

The remainder of this paper is organized as follows. Section~\ref{sec:background} provides background information about the sulphonation process at the factory. Section~\ref{sec:dataset}
presents a description of the utilised dataset. In section~\ref{sec:experiments} we describe the experiments and present our results. Finally, in  section~\ref{sec:conclusions} we conclude.

\section{Background}
\label{sec:background}

At Unger Fabrikker AS, the chemical sulphonation process is part of the tasks involved to manufacture active detergents. The sulphonation reaction is based on several sulphonation reagents~\cite{ref1}. This process consists of burning Sulphur and the conversion of SO\textsubscript{2} gas to SO\textsubscript{3} gas. The SO\textsubscript{3} gas is diluted with air and mixed with organic liquid (a fatty acid raw material) in a liquid-gas reactor. One of the key elements to the burning of Sulphur and the conversion of SO\textsubscript{2} gas to SO\textsubscript{3} gas is the dew point of the air within the reactor. The result from the reactor is a sulphonic acid with different qualities that depend on the type of organic liquid adopted during the sulphonation phase.

In order to measure the quality of the reaction, an NT-value is computed. This is determined by how much of the organic liquid is sulphonated, and this is expressed by a neutralization-number (NT). In plain chemistry, it defines how many milligrams of Kalium Hydroxid (KOH) one needs to neutralize one gram of sulphonic acid~\cite{ref2}. To ascertain the neutralization number, the Karl Fischer's titration method is used~\cite{ref3}.

There are several transitions during one week in the chemical production line at Unger, where they switch between producing a variety of different outputs. Hence the neutralization number will differ with respect to what specific output they are producing at any given time. To control the performance of the shift in real-time, in terms of the product quality (NT), there is a great need for feasible real-time product analysis and monitoring. Today, results from manual laboratory analysis have a delay of approximately $30$ minutes. The effect of this, is that the production will be on hold causing a costly delay. To overcome this, Unger aspires to use machine learning trained models to predict NT, in order to provide the operator with continuous estimates. This has the potential to reduce waste, by not producing unqualified output when the process is in a ``blind spot'', and furthermore lead to cost saving and reduced operator exposure, by reducing the amount of manual sampling and laboratory testing.

\section{Dataset Information}
\label{sec:dataset}

During the process, samples from the production line are taken by an specialist. Then, those samples are analyzed using the tritration technique. Based on the computed results, the specialist tunes different parameters of the process in order to achieve the desired quality standard. The task of taking the sample and obtaining the final results requires around 30 minutes. A database is used to store the analysis results.

This allows further inspection and analysis. The typical process parameters in a sulphonation process are temperature, flow,  pressure and potential of hydrogen (pH). In this case, the prediction of the NT-value was based on the following $8$ process variables listed in Table~\ref{tab:parameters}.

\begin{table}[ht!]
\centering
 \caption{Chemical process variables.}
    \begin{tabular}{C{2cm}|C{4cm}}
    \textbf{Variable}    & \textbf{ID} \\ \hline
    Raw-material. & The quantity of organic material in kg/hr. \\ \hline
    Sulfur & The quantity of sulfur in kg/hr. \\ \hline
    Dew-point & The value of how dry the air is. Measured in temperature. \\ \hline
    Air-sulfur-oven & This is the quantity of air injected into the sulfur oven nm\textsuperscript{3}/hr. \\ \hline
    Air-converter & Quantity of air added into the converter in nm\textsuperscript{3}/hr. \\ \hline
    Air-SO3-filter & The amount of air injected into the SO3 filter in nm\textsuperscript{3}/hr. \\ \hline
    Molar & The mol rate. \\ \hline
    Molar-stp & The molar weight. \\ \hline
    \end{tabular}
    \label{tab:parameters}
\end{table}

In order to make the plots more readable, each variable was given a shorter id. Table~\ref{tab:mapping} shows the corresponding ids.

Because of confidentiality purposes, all the parameters were normalized. This was done by subtracting the mean and dividing by the standard deviation. Overall, the database has $14,252$ sample points from which $23$ were outliers. The outliers correspond to erroneous values in one or more of the variables. Currently, the erroneous measurements are identified and catalogued by an experienced engineer. In this work, we conducted the experiments after outliers were removed. The automatic detection of possible outlier values will be left as future work.

\begin{table}[ht!]
\centering
 \caption{Chemical process variables and their IDs.}
    \begin{tabular}{cc}
    \textbf{Variable}    & \textbf{ID} \\ \hline
    Raw-material & A \\ \hline
    Sulfur & B \\ \hline
    Dew-point & C \\ \hline
    Air-sulfur-oven & D \\ \hline
    Air-converter & E \\ \hline
    Air-SO3-filter & F \\ \hline
    Molar & G \\ \hline
    Molar-stp & H \\ \hline
    NT & NT \\ \hline
    \end{tabular}
    \label{tab:mapping}
\end{table}

\section{Experiments and results}
\label{sec:experiments}

As previously stated, the objective is to predict the NT value based on the aforementioned 8 variables, and to identify which of those are the most relevant. To this extent, we trained different predictive models. Specifically, a linear regression model a regression tree and a Random Forest (with 100 trees). For the regression tree, we used the rpart R library~\cite{rpart,rpart.plotpackage}. To avoid overfitting, the dataset was randomly divided into \emph{training}, \emph{validation} and \emph{testing} sets. The training set corresponds to 80\% of the total data. The remaining 20\% is used as the test set. The validation set was generated with 20\% from the train set.

First, we conducted a feature importance analysis. The aim of the feature analysis was to find the most important features, i.e., the ones that provide the strongest predictive power. Then, based on this, we trained several predictive models with different subsets of features in order to assess how the prediction error varied.

\subsection{Feature importance analysis.}

\begin{figure}[ht!]
\centering
\includegraphics[width=1.0\columnwidth]{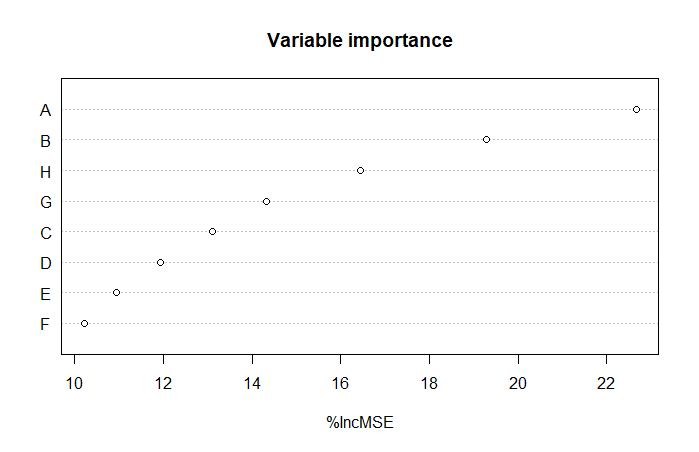}
\caption{Variable importance plot from Random Forest model.}
\label{fig:importance}
\end{figure}

To obtain the feature importance ranking, we trained a Random Forest model on the training set using the R library \emph{randomForest}~\cite{randomForestLib}. This library includes a method to rank the most relevant features based on the out-of-bag error. The error was determined by the percent increase of the mean standard error (MSE), which was computed with the out-of-bag data points as input. The prediction error from the out-of-bag samples is estimated for every tree. Afterwards, the same procedure is conducted after permuting every feature and the differences are averaged across the trees. Then, they are  normalized using the standard deviation of the differences. Figure~\ref{fig:importance} shows the resulting feature ranking with the most important variables at the top. The increase of the mean standard error is shown in the $x$ axis as a percent. That is, what is the expected error increase if that variable is removed. According to this, the most important feature is A (Raw-material), followed by B (Sulfur) and so forth. Here, it can be seen that the difference between H and G is big, while the differences between the lower ranked variables are smaller.

\begin{figure}[ht!]
\centering
\includegraphics[width=1.0\columnwidth]{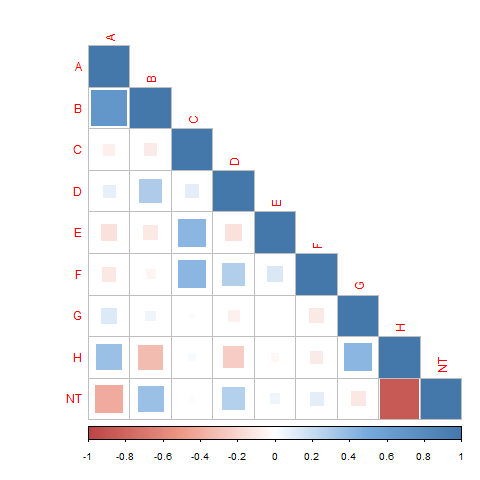}
\caption{Correlation plot.}
\label{fig:corr}
\end{figure}

In order to further explore the relation between the most important variables and the response (NT), we generated a correlation plot (Figure~\ref{fig:corr}). This plot shows the Pearson correlation between each pair of variables. The squares' color and size are proportional to the correlation coefficients. Here, we can see that NT has a high negative correlation with A and H, and a high positive correlation with B, which are also the most important variables according to the Random Forest results.

\subsection{Forward feature selection.}

On the foundation of the previous feature importance analysis, we then trained each model starting with just the most important variable. In this case A. Next, we added the second most important variable, and so forth. The models were then trained and validated. Table~\ref{tab:results_lr} shows the results of the linear regression model for the $8$ configurations, starting with a single feature and adding one by one in order of importance. The table shows the mean absolute error (MAE), the root mean squared error (RMSE) and the Pearson correlation (corr) \cite{paerson}. The final column shows the features used to train the model.

\begin{table}[ht]
\centering
\caption{Results with linear regression.}
\begin{tabular}{rrrrl}
  \hline
 & RMSE & MAE & corr & features \\ 
  \hline
1 & 0.89678 & 0.75812 & 0.40166 & A \\ 
  2 & 0.25392 & 0.12088 & 0.96578 & A,B \\ 
  3 & 0.25323 & 0.12040 & 0.96597 & A,B,H \\ 
  4 & 0.25315 & 0.11963 & 0.96598 & A,B,H,G \\ 
  5 & 0.25304 & 0.11955 & 0.96602 & A,B,H,G,C \\ 
  6 & 0.25285 & \underline{0.11923} & 0.96607 & A,B,H,G,C,D \\
  7 & 0.25289 & 0.11940 & 0.96606 & A,B,H,G,C,D,E \\ 
  8 & \underline{0.25223} & 0.11945 & \underline{0.96623} & A,B,H,G,C,D,E,F \\
   \hline
\end{tabular}
\label{tab:results_lr}
\end{table}

\begin{table}[ht]
\centering
\caption{Results with regression tree.}
\begin{tabular}{rrrrl}
  \hline
 & RMSE & MAE & corr & features \\ 
  \hline
1 & 0.54544 & 0.31468 & 0.83157 & A \\ 
  2 & \underline{0.28547} & \underline{0.13773} & \underline{0.95671} & A,B \\ 
  3 & 0.34289 & 0.15580 & 0.93666 & A,B,H \\ 
  4 & 0.34289 & 0.15580 & 0.93666 & A,B,H,G \\ 
  5 & 0.34289 & 0.15580 & 0.93666 & A,B,H,G,C \\ 
  6 & 0.34289 & 0.15580 & 0.93666 & A,B,H,G,C,D \\ 
  7 & 0.34289 & 0.15580 & 0.93666 & A,B,H,G,C,D,E \\ 
  8 & 0.34289 & 0.15580 & 0.93666 & A,B,H,G,C,D,E,F \\
   \hline
\end{tabular}
\label{tab:results_t}
\end{table}

\begin{table}[ht]
\centering
\caption{Results with Random Forest.}
\begin{tabular}{rrrrl}
  \hline
 & RMSE & MAE & corr & features \\
  \hline
1 & 0.58608 & 0.31142 & 0.81099 & A \\ 
  2 & 0.23970 & 0.10459 & 0.96970 & A,B \\ 
  3 & 0.23250 & 0.09997 & 0.97139 & A,B,H \\ 
  4 & 0.23041 & 0.09870 & 0.97190 & A,B,H,G \\ 
  5 & 0.22979 & 0.09825 & 0.97207 & A,B,H,G,C \\ 
  6 & 0.22700 & 0.09288 & 0.97274 & A,B,H,G,C,D \\ 
  7 & 0.22492 & 0.09165 & 0.97324 & A,B,H,G,C,D,E \\ 
  8 & \underline{0.22258} & \underline{0.08998} & \underline{0.97381} & A,B,H,G,C,D,E,F \\ 
   \hline
\end{tabular}
\label{tab:results_rf}
\end{table}

\begin{figure*}[ht!]%
    \centering
    \subfloat{{\includegraphics[width=8.5cm]{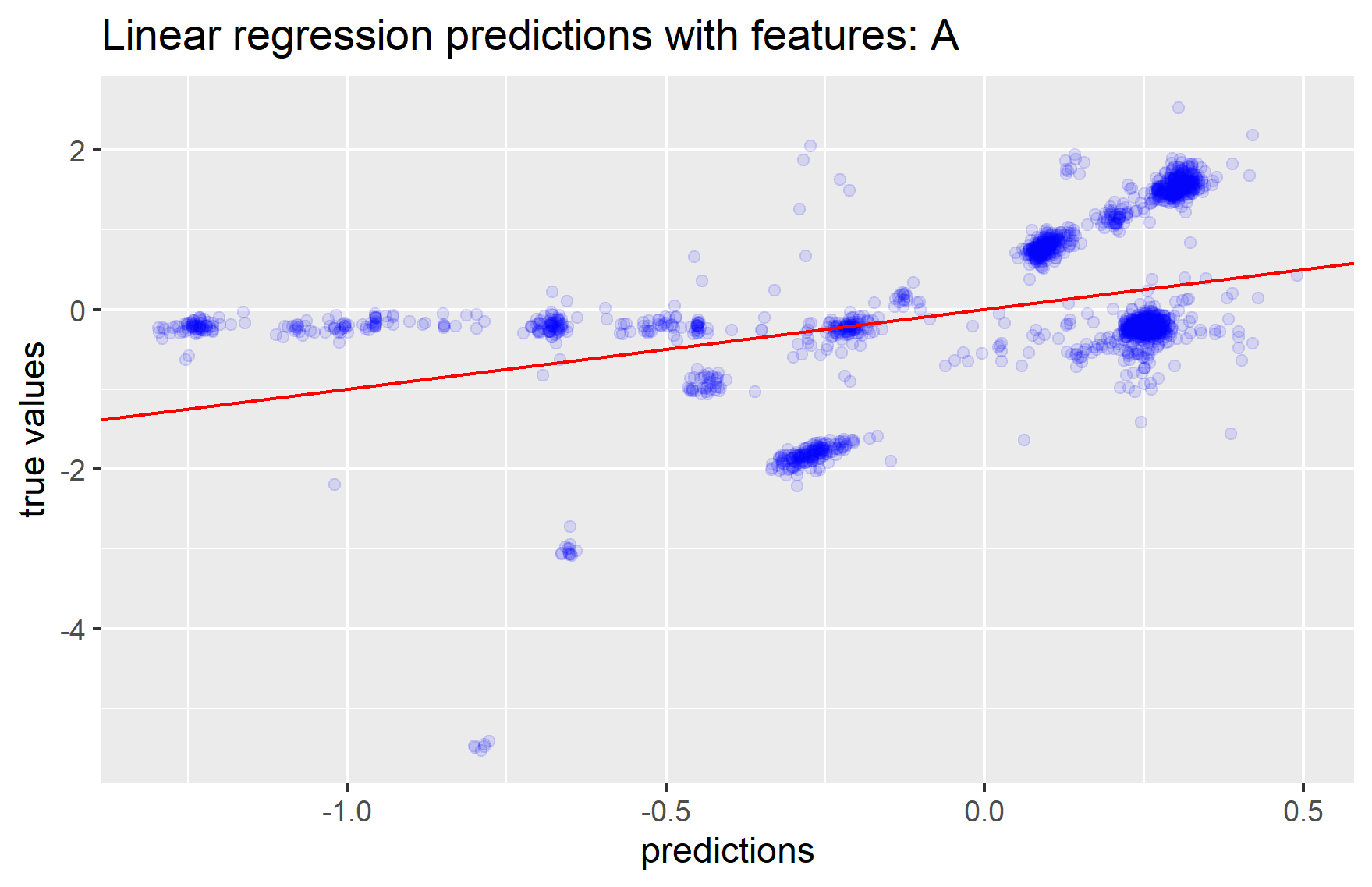} }}%
    \qquad
    \subfloat{{\includegraphics[width=8.5cm]{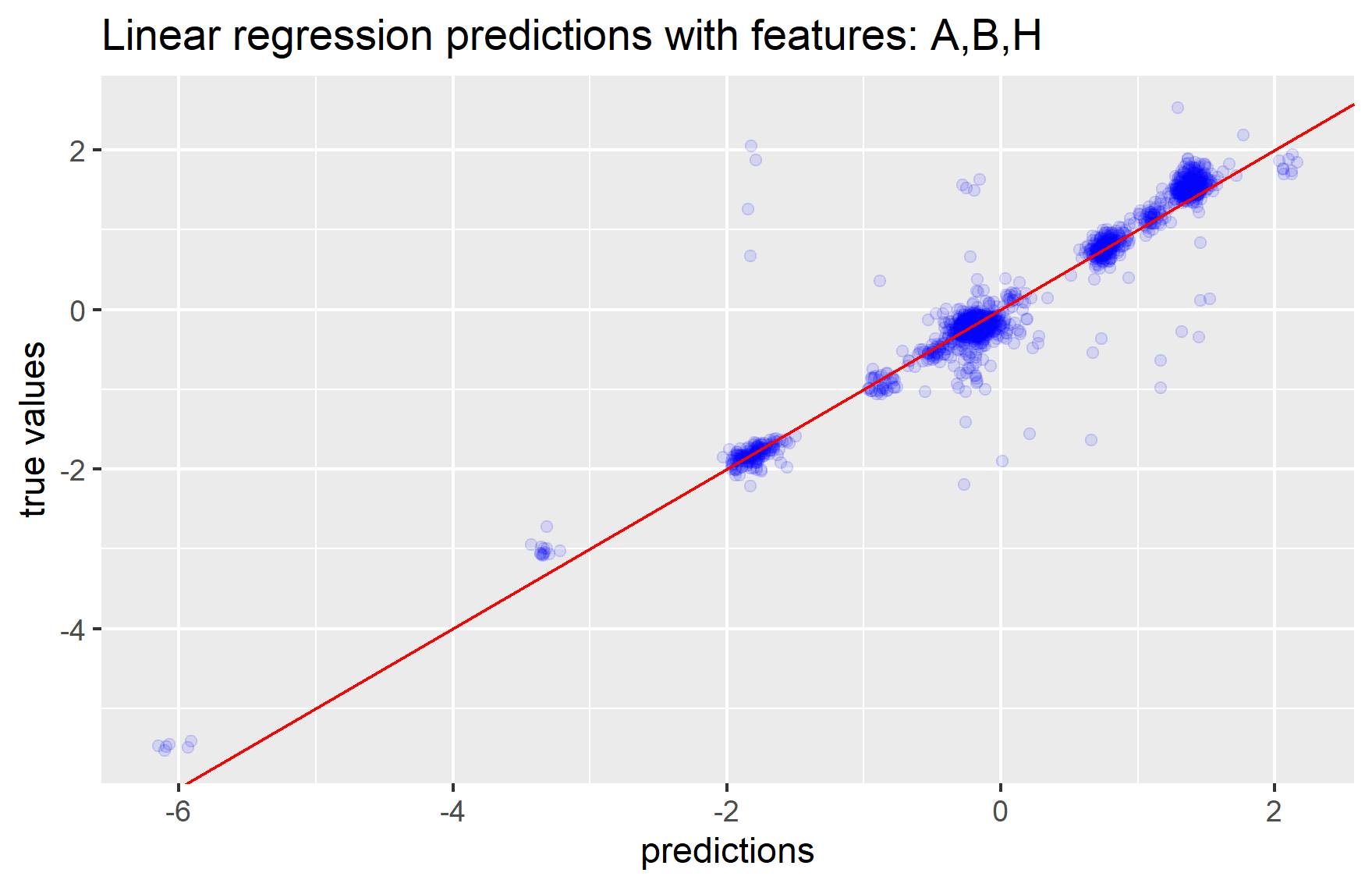} }}%
    \caption{Predictions v.s. true values with linear regression. Using feature A (left) and using features A,B and H (right).}%
    \label{fig:lr_predictions}%
\end{figure*}

Here, we can see that the RMSE when using just A is $0.896$ and it is reduced to $0.253$ when adding B, a reduction of $71.0\%$ of the error. The lowest RMSE was achieved when using all features however it is not very different when using, e.g., the top 3 features. The highest correlation was also achieved with all features. This preliminary results suggest that features C,D,E and F do not provide any extra benefit for linear regression, however more extensive experiments would be required to make a strong conclusion but we believe this is a good estimation.

Similarly, Tables~\ref{tab:results_t} and ~\ref{tab:results_rf} show the results for regression tree and Random Forest, respectively. For regression tree, the optimal was achieved when using features A and B. For Random Forest, the lowest error was obtained when using all variables. Overall, Random Forest achieved the best results. This was expected since in general, ensemble methods perform better than more simple individual models~\cite{polikar2012ensemble}.

Figure~\ref{fig:lr_predictions} shows the predictions v.s. the true values for the linear regression model. Here, we can see that the predictions are more accurate when using features A,B,H as compared to only using A. For all models, the performance when using just a single variable (A) was very poor but it drastically improved when adding variable B.

\begin{figure}[ht!]
\centering
\includegraphics[width=1.0\columnwidth]{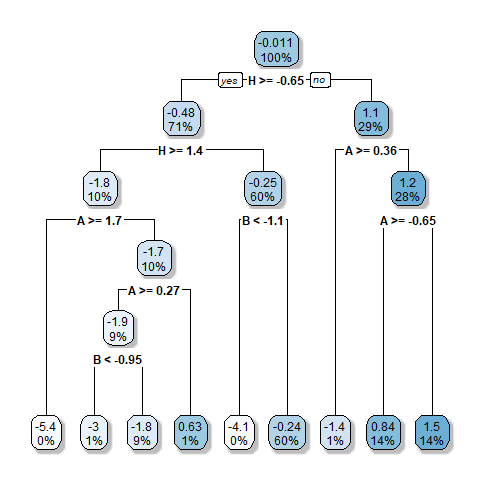}
\caption{Resulting regression tree when using all features. However, the tree algorithm only chose A, B and H.}
\label{fig:tree}
\end{figure}

One thing to note is that the regression tree produces the same results after adding feature H. To understand why, we plotted the resulting trees and it turned out that all of them were the same regardless of the number of selected features. This is because the tree algorithm only chose the variables A,B and H. Figure~\ref{fig:tree} shows the resulting tree when using all features (but just A,B and H were selected by the tree algorithm). Every node shows the predicted value and the percent of samples contained in a given node. This provides more evidence that features A,B and H are important since those were also chosen as the top ones by the Random Forest variable importance analysis.

Additionally, we evaluated feature importance with filter methods \cite{BOMMERT2020106839} namely: chi-squared, gain-ratio and correlation. Table~\ref{tab:filter_methods} shows the resulting variable importance (descending) for each method. From this table we confirm that the most important variables are H, A and B. The three methods also agree that C is the least important variable which makes sense since from Figure~\ref{fig:corr} it can be seen that its correlation with the response variable NT is close to $0$. From these tests we can observe that \emph{Raw-material}, \emph{Sulfur} and \emph{Molar-stp} are the most relevant variables when predicting NT.

\begin{table}[ht!]
\centering
 \caption{Variable importance (descending order) with different filter methods.}
    \begin{tabular}{cc}
    \textbf{Method}    & \textbf{Importance} \\ \hline
    chi-squared & H, A, B, G, E, F, D, C \\ \hline
    gain ratio & H, B, A, G, E, D, F, C \\ \hline
    correlation & H, A, B, D, G, F, E, C \\ \hline
    \end{tabular}
    \label{tab:filter_methods}
\end{table}

\subsection{Results on the test set.}

Now that we have identified \emph{Raw-material}, \emph{Sulfur} and \emph{Molar-stp} as the most important variables, we evaluate the regression models on the test set only using those three variables. Table~\ref{tab:results_testset} shows the results on the test set.

\begin{table}[ht]
\centering
\caption{Results with linear regression.}
\begin{tabular}{rrrrl}
  \hline
 Method & RMSE & MAE & corr \\ 
  \hline
  Linear regression & 0.24522 & 0.1197828 & 0.9695145 \\ 
  Regression tree & 0.3352072 & 0.1531236 & 0.9421266 \\
  Random Forest & 0.2253893 & 0.1010736 & 0.9743675 \\ 
  \hline
\end{tabular}
\label{tab:results_testset}
\end{table}

Again, Random Forest performed the best by only using the 3 most important variables found during the previous experiments. These results are similar to the ones using the complete set of variables. For an industrial process like this, understanding the contributions of each variable is of great importance. For example, if the \emph{Air-So3-filter} parameter is producing inconsistent values, the remaining variables could still be used to predict the NT with acceptable accuracy. This analysis is one of the first steps towards building a robust system that can automate the sulphonation process. For example, a system could be in charge of monitoring the consistency and quality of the parameter values. If the quality of a given set of parameters is not within the limits then, combinations of the remaining parameters could still be used to predict the NT value without having to interrupt the process.

\section{Conclusions}
\label{sec:conclusions}

In this paper we have analyzed a dataset of process variables, collected from an industrial chemical sulphonation process. The task was to predict the product quality determining parameter (NT). As the NT is infeasible to measure directly in real-time, we used soft sensing to predict it and we assessed the process variable's relevance. To this extent, we trained three different regression models: linear regression, regression tree and Random Forest. We also conducted a feature importance analysis. Our results show that it was possible to achieve good predictions using only $3$ (Raw-material, Sulfur, Molar-stp) out of the $8$ variables. We also found that the regression tree model only chose three variables (Raw-material, Sulfur, Molar-stp), which happened to be identical to the $3$ most important identified by the out-of-bag error of random forest and the three applied filter-based feature weighting methods. One of the limitations of this work is that we used a forward feature selection approach for some of the analyses. Despite its computational efficiency, it also leaves out several possible variable combinations. A more exhaustive selection method will be considered as future work.

This analysis serves as a good starting point and as baseline to further develop an application to lower the prediction error on this specific dataset. Hence, this  is a first step towards the implementation of a fully automated sulphonation process, with the final goal of reducing time gaps between product transitions, and ultimately production waste and human exposure to acidic environments.

\section*{ACKNOWLEDGMENT}

Research leading to these results has received funding from the EU ECSEL Joint Undertaking under grant
agreement no 737459 (project Productive4.0) and from the Research Council of Norway.

\bibliographystyle{IEEEtran}
\bibliography{IEEEabrv,references}

\end{document}